\pdfoutput=1

\documentclass[11pt]{article}

\usepackage[final]{acl}

\usepackage{times}
\usepackage{latexsym}
\usepackage[T1]{fontenc}

\usepackage[utf8]{inputenc}

\usepackage{microtype}

\usepackage{inconsolata}

\usepackage{graphicx}

\usepackage{booktabs}
\usepackage{tabularx}
\usepackage{cleveref}

\title{HITSZ's End-To-End Speech Translation Systems Combining Sequence-to-Sequence Auto Speech Recognition Model and Indic Large Language Model for IWSLT 2025 in Indic Track}

\author{
 \textbf{Xuchen Wei},
 \textbf{Yangxin Wu},
 \textbf{Yaoyin Zhang},
 \textbf{Henglyu Liu},
\\
 \textbf{Kehai Chen\textsuperscript{*}},
 \textbf{Xuefeng Bai},
 \textbf{Min Zhang}
\\
 School of Computer Science and Technology, Harbin Institute of Technology, Shenzhen, China
\\
 \texttt{\{2023311524,2023311526,2023313720,23s151043\}@stu.hit.edu.cn, } \\
 \texttt{\{chenkehai,baixuefeng,zhangmin2021\}@hit.edu.cn} \\
}

\begin{document}

\maketitle
\begingroup
\renewcommand\thefootnote{}\footnote{\textsuperscript{*}Corresponding author}
\addtocounter{footnote}{-1}
\endgroup

\begin{abstract}
This paper presents HITSZ's submission for the IWSLT 2025 Indic track, focusing on speech-to-text translation (ST) for English-to-Indic and Indic-to-English language pairs. To enhance translation quality in this low-resource scenario, we propose an end-to-end system integrating the pre-trained Whisper automated speech recognition (ASR) model with Krutrim, an Indic-specialized large language model (LLM). Experimental results demonstrate that our end-to-end system achieved average BLEU scores of $28.88$ for English-to-Indic directions and $27.86$ for Indic-to-English directions. Furthermore, we investigated the Chain-of-Thought (CoT) method. While this method showed potential for significant translation quality improvements on successfully parsed outputs (e.g. a $13.84$ BLEU increase for Tamil-to-English), we observed challenges in ensuring the model consistently adheres to the required CoT output format.
\end{abstract}

\section{Introduction}

Speech-to-text translation plays a vital role in overcoming language barriers in multilingual and international contexts, such as real-time translation during online meetings. Although translation systems for high-resource language pairs have achieved impressive performance, low-resource language pairs, particularly those involving Indic languages, continue to face significant challenges \citep{radford_robust_2023, joshi2020state}.

\begin{figure*} 
\vspace{-2.2cm}
\centering
\includegraphics[width=\textwidth]{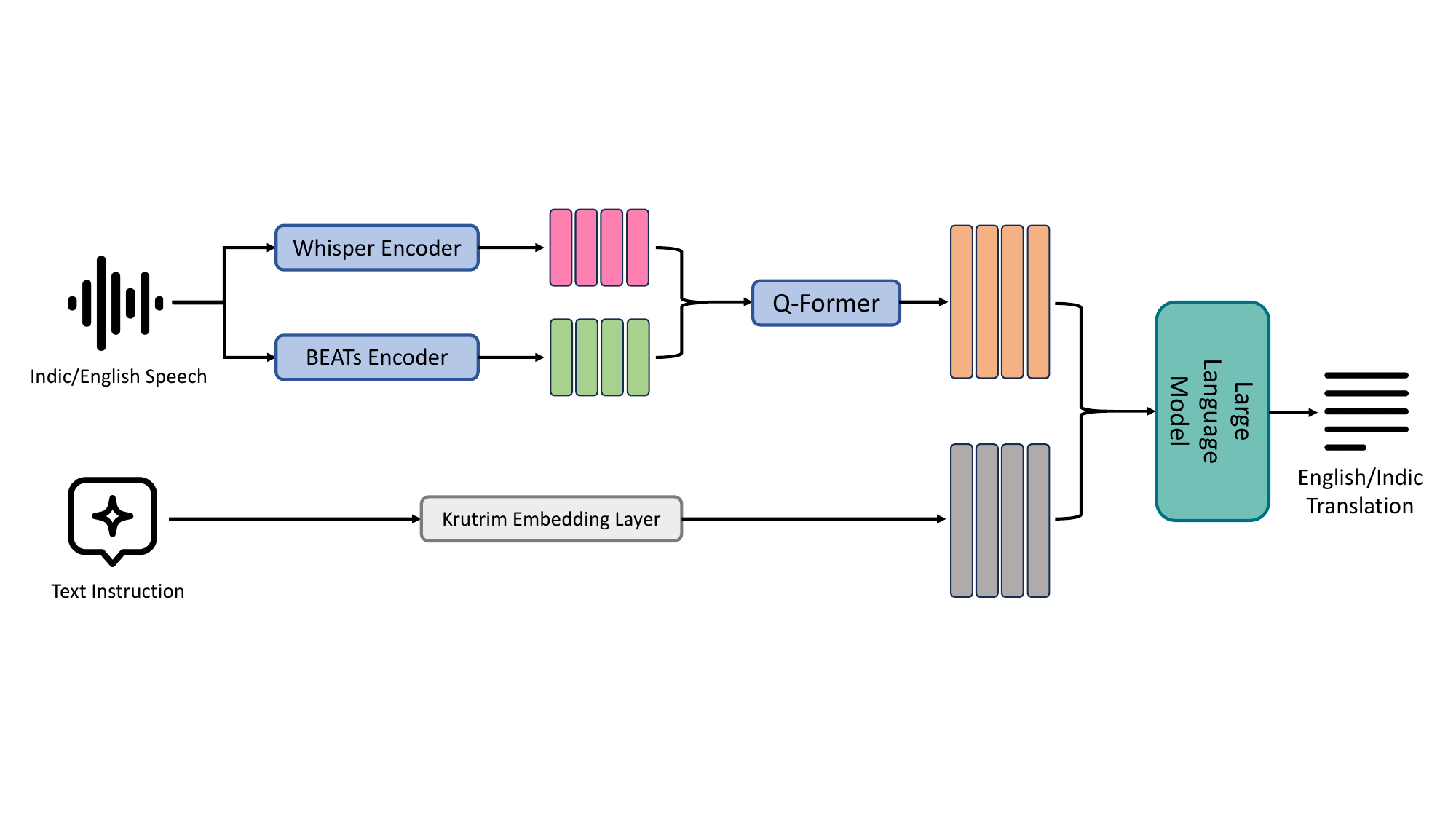}
\vspace{-3cm}
\caption{Overview of the our end-to-end spoken language translation system.}
\label{fig:frame}
\end{figure*}

This paper presents HITSZ’s submission to the Indic Track of IWSLT 2025, covering bidirectional speech translation between English and three major Indic languages: Hindi, Bengali, and Tamil. An overview of the end-to-end system is illustrated in \Cref{fig:frame}.

Data scarcity poses a significant challenge for speech translation (ST) between English and Indic languages, primarily due to the low-resource nature of these language pairs and the reliance on data-driven neural models \citep{ahmad-etal-2024-findings}. Acknowledging this, we collected available parallel corpus from the official IWSLT data releases for effective end-to-end ST model training.

Cascade and end-to-end (E2E) systems represent two prominent paradigms in ST, each offering distinct advantages \citep{ney1999speech, mathias2006statistical, berard2016listen}. While cascaded systems typically achieve higher translation quality \citep{agarwal2023findings}, E2E systems are favored for their lower latency and reduced modeling complexity \citep{ahmad-etal-2024-findings, xu2023recent}. This work focuses exclusively on the end-to-end paradigm for the bidirectional speech translation task. We adopt an \textit{unconstrained} setting and utilize state-of-the-art pre-trained models, including Whisper \citep{radford_robust_2023} and Krutrim \citep{kallappa_krutrim_2025}, to develop E2E systems for both English-to-Indic and Indic-to-English directions. Although additional resources such as the IndicVoices \citep{javed2024indicvoicesbuildinginclusivemultilingual} dataset are available, we deliberately exclude them due to concerns about potential overlap with the test set.

The remainder of this paper is structured as follows: \Cref{Related Work} reviews related work on speech translation, particularly in low-resource and Indic language settings. \Cref{Data} describes the datasets and data pre-processing. \Cref{Method} introduces our end-to-end system. \Cref{Experiments and Results} presents the experimental settings, results, and analysis. Lastly, \Cref{Conclusion} concludes the paper.

\section{Related Work}
\label{Related Work}
Recent advances in end-to-end speech translation (ST) have demonstrated the effectiveness of combining large pre-trained models with task-specific adaptation \citep{wang-etal-2017-instance, berard2018end, bansal2019pre, wang2020fairseq, alinejad2020effectively}, especially in low-resource and multilingual settings \citep{marie-etal-2019-nicts,sun-etal-2020-knowledge-distillation,tsiamas_pushing_2024,li-etal-2025-mit}. Among them, several works stand out for their innovative training paradigms and architectural choices that have directly influenced our approach. These include NICT’s submission to IWSLT 2024 \citep{dabre-song-2024-nicts}, which leverages decoder-side fine-tuning of Whisper with pseudo-labels from IndicTrans2 \citep{gala_indictrans2_2023}; ZeroSwot, which introduces an encoder-centric alignment method for zero-shot ST \citep{tsiamas_pushing_2024}; and SALMONN, a multimodal framework that uses a lightweight training pipeline to adapt frozen encoders and LLMs through cross-modal instruction tuning \citep{tang2024salmonn}. In what follows, we briefly review each of these works and highlight their relevance to our system design.

\subsection{NICT’s E2E ST System in IWSLT 2024}
One of the most relevant works to our approach is the IWSLT 2024 submission by NICT, which developed end-to-end speech translation systems for English to Hindi, Bengali, and Tamil. A key contribution was their fine-tuning strategy for Whisper: instead of using human-annotated translations, they first fine-tuned IndicTrans2 to generate pseudo-translations from English transcripts. These synthetic targets were then used to train Whisper in a speech-to-text translation setting, effectively distilling knowledge and improving decoder performance beyond what reference translations alone could achieve.

\subsection{ZeroSwot}
Another influential work is ZeroSwot, which proposes a novel zero-shot end-to-end ST framework by aligning speech representations with the embedding space of a multilingual MT model. In their setup, the speech encoder is initialized from a CTC-finetuned wav2vec 2.0 model \cite{baevski_wav2vec_2020} and trained using a combination of CTC loss and Optimal Transport loss \citep{graves2006connectionist, peyre2019computational}. The goal is to produce subword-level acoustic representations that match those expected by a frozen multilingual MT encoder (NLLB) \citep{nllbteam2022languageleftbehindscaling}. In addition, a compression adapter \citep{liu2020bridging} is introduced to map variable-length audio sequences into subword-aligned embeddings, bridging both length and representation mismatches between modalities. In contrast to NICT’s decoder-focused fine-tuning, ZeroSwot emphasizes encoder-side alignment, enabling zero-shot translation without requiring any parallel ST data.

\subsection{SALMONN}
We also take inspiration from SALMONN, a multimodal framework that integrates Whisper and BEATs \citep{chen_beats_2023} encoders with a large language model (Vicuna) \citep{vicuna2023} to enable general auditory understanding across speech, audio events, and music. Although SALMONN targets a broader set of audio-language tasks beyond ST, its modular design and training strategy are particularly relevant. SALMONN adopts a three-stage training pipeline—pre-training \citep{zhu-etal-2024-towards-robust}, instruction tuning, and activation tuning—where only lightweight modules (Q-Former and LoRA adaptors \citep{li_blip-2_2023, hu2022lora}) are updated while the encoders and LLM remain frozen. This design enables efficient adaptation with minimal parameter updates. Our work builds on this principle by leveraging pre-trained components and applying modular fine-tuning in a similarly efficient manner, tailored to low-resource, bidirectional speech translation between English and Indic languages.

\section{Data}
\label{Data}

In this section, we present the statistics of the initial corpora and describe our methods for pre-processing the raw data.

\subsection{Dataset}

\begin{table}[h]
  \centering
  \resizebox{\columnwidth}{!}{
    \begin{tabular}{lcccc}
      \toprule
      \textbf{Direction} & \textbf{Train} & \textbf{Dev} & \textbf{Test} &
      \begin{tabular}[c]{@{}c@{}} \textbf{Total} \\ \textbf{Speech Hours} \end{tabular} \\
      \midrule
      en $\rightarrow$ bn & 680.9  & 40.8  & 93.2  & 814.9 \\
      en $\rightarrow$ hi & 680.9  & 40.8  & 93.2  & 814.9 \\
      en $\rightarrow$ ta & 680.9  & 40.8  & 93.2  & 814.9 \\
      bn $\rightarrow$ en & 158.0  & 1.0  & 1.3  & 160.3 \\
      hi $\rightarrow$ en & 653.9  & 1.0  & 1.3  & 656.2 \\
      ta $\rightarrow$ en & 478.2  & 1.0  & 2.2  & 481.4 \\
      \bottomrule
    \end{tabular}
  }
  \caption{Statistics of dataset for training, development, and test sets. The abbreviations \textit{en}, \textit{bn}, \textit{hi}, and \textit{ta} stand for English, Bengali, Hindi, and Tamil, respectively.}
  \label{tab:data_stats}
\end{table}

We rely solely on the corpus provided by the organizers, with its statistics detailed above. Although we did not incorporate any supplementary data, our model remains \textit{unconstrained} by leveraging the pre-trained Whisper ASR model, the Krutrim large language model, and adapters trained specifically for the spoken language translation task based on these two models. The audio segments corresponding to textual sentences are extracted from the original files based on the given offset and duration details. Post-segmentation, each dataset entry includes an audio clip in the source language, its transcription, and a translation in the target languages.

\subsection{Pre-processing}

We find that some audio clips in the English-to-Indic corpus are very long, indicating a very large consumption of GPU memory. To accelerate the fine-tuning process, we separate the data with English transcription length less than and above 400 characters, which allows us to increase the batch size during the training process.

\section{Method}
\label{Method}

Our model builds upon the \textit{Dhwani} model \citep{sanket2025IndicST}, which is trained for speech translation tasks in Indic languages and is itself derived from the SALMONN architecture. To effectively process and align multimodal audio data with textual outputs, the architecture integrates several specialized components. For speech signals, it leverages the Whisper speech encoder (WhisperSE) to extract robust linguistic representations. In parallel, non-speech audio inputs, such as environmental sounds and music, are processed using the BEATs encoder, which is optimized for general audio understanding.

These two audio streams are subsequently bridged to the language model via a Window-Level Query Transformer (Q-Former), which acts as a connection module to transform modality-specific features into a unified representation space. The transformed tokens are then passed to the Krutrim LLM, a 7-billion-parameter dense transformer model built on a multilingual foundation and optimized for Indic language tasks. Trained on a corpus of 2 trillion tokens with extensive coverage of native Indian languages, Krutrim demonstrates strong performance across multilingual benchmarks in both Indic and English, despite being relatively lightweight in terms of training compute.

To enable efficient domain-specific adaptation without retraining the entire model, Low-Rank Adaptation (LoRA) is employed during fine-tuning. This technique aligns the LLM’s outputs with the semantics of the input audio, facilitating robust and adaptable performance.

\section{Experiments and Results}
\label{Experiments and Results}

This section details the experimental setup and presents the results for our monolingual speech translation models, trained individually for each translation direction. We follow the settings of the \textit{Dhwani} model, employing the \textit{Whisper-large-v2} model as the speech encoder and the \textit{Krutrim-1-instruct} model as the text decoder branch. 

\subsection{English-Indic Translation}

For the English-to-Indic translation task, we adopted a fine-tuning strategy where both the WhisperSE and the BEATs audio encoders were kept frozen. Training focused exclusively on the Q-Former module, which connects the speech encoder to the language model, and a LoRA adapter integrated into the LLM branch. We configured the LoRA adapter with a rank ($r$) of 8 and an alpha ($\alpha$) of 32.

The learning rate schedule commenced with a linear warmup phase over the initial 3,000 training steps, increasing from a base rate of $1e^{-6}$ to the peak learning rate of $3e^{-5}$. Subsequently, the learning rate followed a cosine decay schedule, oscillating between the maximum rate ($3e^{-5}$) and a minimum rate ($1e^{-5}$), before finally decaying to the minimum rate of $1e^{-5}$.

\begin{table}[h]
  \centering
    \begin{tabular}{lcc}
      \toprule
      \textbf{Direction} & \textbf{Dev} & \textbf{Test} \\
      \midrule
      en $\rightarrow$ bn & 30.61 & 27.00 \\
      en $\rightarrow$ hi & 37.83 & 33.84 \\
      en $\rightarrow$ ta & 25.97 & 22.81 \\
      \bottomrule
    \end{tabular}
  \caption{BLEU scores on the development and test set in English-to-Indic directions.}
  \label{tab:en-indic_results}
\end{table}

We initiated training using only the \textit{short} audio segments from our dataset. This allowed for a larger batch size of 4, thereby accelerating the training process. Models were trained independently for three language pairs: English-to-Bengali, English-to-Hindi, and English-to-Tamil. For each pair, the checkpoint yielding the highest BLEU score on the development set was selected for subsequent incremental fine-tuning on the dataset containing \textit{long} audio segments. Detailed results are presented in \Cref{tab:en-indic_results}.

\subsection{Indic-to-English Translation}

The experimental setup for Indic-to-English translation largely mirrored the English-to-Indic configuration. However, a key difference was the absence of exceptionally \textit{long} audio clips in the Indic-to-English corpus. Consequently, we did not employ the two-stage (\textit{short/long}) training strategy used for the English-to-Indic directions.

Recognizing that the Whisper model exhibits comparatively lower performance on Indic languages than high-resource languages, we set the WhisperSE module to be trainable for the first epoch. Using a batch size of 1 with gradient accumulation over 4 steps helped conserve GPU memory while enabling updates to the WhisperSE, aiming for improved feature extraction from Indic audio inputs. The evaluation results are presented in \Cref{tab:indic-en_results}.

To mitigate the challenge of limited training data and to better exploit the inherent bilingual capabilities of the LLM, we explored the Chain-of-Thought (CoT) prompting and fine-tuning technique.  Specifically, this approach involved fine-tuning the model to first produce a transcription of the speech in the source language, followed by the English translation. Our findings indicate that the automatic parsing of the generated responses for the reliable extraction of the final translation output was not consistently successful.

Our experiments on the development set demonstrate that, on average, $66.54\%$ of the responses generated by our E2E system adhere to the Chain-of-Thought (CoT) format constraints and can be successfully parsed. For the subset of responses that are parsable, results indicate notable improvements in BLEU scores. Specifically, in the Tamil-to-English translation direction, we observe a significant BLEU score improvement of 13.84 points.

\begin{table}[h]
  \centering
    \begin{tabular}{lccc}
      \toprule
      \textbf{Direction} & \textbf{Dev} & \textbf{Test} \\
      \midrule
      bn $\rightarrow$ en & 25.38 & 25.02 \\
      hi $\rightarrow$ en & 31.71 & 39.29 \\
      ta $\rightarrow$ en & 20.93 & 19.27 \\
      \bottomrule
    \end{tabular}
  \caption{BLEU scores on the development and test set in Indic-to-English directions.}
  \label{tab:indic-en_results}
\end{table}

\begin{table}[h]
  \centering
  \resizebox{\columnwidth}{!}{
    \begin{tabular}{lccc}
      \toprule
      \textbf{Direction} & \begin{tabular}[c]{@{}c@{}} \textbf{CoT Parsing} \\ \textbf{Success Rate} \end{tabular} & \textbf{BLEU Score} & \textbf{$\Delta$} \\
      \midrule
      bn $\rightarrow$ en & 68.18\% & 28.13 & 2.592 \\
      hi $\rightarrow$ en & 71.00\% & 38.49 & 6.780 \\
      ta $\rightarrow$ en & 60.43\% & 34.77 & 13.84 \\
      \bottomrule
    \end{tabular}
  }
  \caption{Parsing success rate of Chain of Thought responses in Indic-to-English directions; BLEU scores of successfully parsed CoT responses on the development set; and the corresponding BLEU score improvements of the CoT method.}
  \label{tab:cot_parsing_success_rate}
\end{table}

\section{Conclusion}
\label{Conclusion}

This paper presented HITSZ's submission to the IWSLT 2025 speech-to-text translation task in the Indic track. We leveraged recent advancements in Indic LLM by integrating the Whisper model and the Krutrim model into our end-to-end system. Future work will primarily focus on two key directions: first, enhancing the instruction-following capability of the specialized LLM for Indic languages to facilitate the development of a Spoken Language Translation system utilizing the Chain-of-Thought (CoT) method; and second, improving its generation capabilities in Indic languages to boost performance in English-to-Indic translation tasks.

\section*{Acknowledgement}

We sincerely thank the shared task organizers for their efforts in designing and coordinating the task, as well as the reviewers for their valuable work that made this technical report possible. This work is supported in part by the National Natural Science Foundation of China (62276077, 62406091, and U23B2055), the Guangdong Basic and Applied Basic Research Foundation (2024A1515011205), the Shenzhen Science and Technology Program (KQTD2024072910215406 and ZDSYS20230626091203008), and Shenzhen College Stability Support Plan (GXWD20220817123150002 and GXWD20220811170358002).

\bibliography{custom}

\end{document}